\pdfoutput=1

\documentclass[11pt]{article}

\usepackage[preprint]{coling}

\usepackage{times}
\usepackage{latexsym}

\usepackage[T1]{fontenc}

\usepackage[utf8]{inputenc}

\usepackage{microtype}

\usepackage{inconsolata}

\usepackage{graphicx}

\usepackage{booktabs}

\title{Human Variability vs. Machine Consistency: A Linguistic Analysis of Texts Generated by Humans and Large Language Models}

\author{Sergio E. Zanotto\textsuperscript{1},
  Segun Aroyehun\textsuperscript{2}\\
  \textsuperscript{1}Department of Linguistics \& Cluster of Excellence ``The Politics of Inequality''\\ 
  University of Konstanz \\
  \texttt{sergio.zanotto@uni-konstanz.de}\\
   \textsuperscript{2}Department of Politics and Public Administration\\ 
   University of Konstanz\\
   \texttt{segun.aroyehun@uni-konstanz.de}
  }

\begin{document}
\maketitle
\begin{abstract}
The rapid advancements in large language models (LLMs) have significantly improved their ability to generate natural language, making texts generated by LLMs increasingly indistinguishable from human-written texts. Recent research has predominantly focused on using LLMs to classify text as either human-written or machine-generated. In our study, we adopt a different approach by profiling texts spanning four domains based on 250 distinct linguistic features. We select the M4 dataset from the Subtask B of SemEval 2024 Task 8. We automatically calculate various linguistic features with the LFTK tool and additionally measure the average syntactic depth, semantic similarity, and emotional content for each document. We then apply a two-dimensional PCA reduction to all the calculated features. Our analyses reveal significant differences between human-written texts and those generated by LLMs, particularly in the variability of these features, which we find to be considerably higher in human-written texts. This discrepancy is especially evident in text genres with less rigid linguistic style constraints. Our findings indicate that humans write texts that are less cognitively demanding, with higher semantic content,
and richer emotional content compared to texts generated by LLMs. These insights underscore the need for incorporating meaningful linguistic features to enhance the understanding of textual outputs of LLMs.
\end{abstract}

\section{Introduction}

The rapid advancements in language models have significantly improved their ability to generate natural language, making machine-generated text (MGT) increasingly indistinguishable from human-written text (HWT). This evolution has highlighted the importance of identifying MGT, a task that has attracted substantial attention in the field of Natural Language Processing (NLP). Consequently, numerous competitions and datasets have emerged to tackle this challenge of Authorship Attribution (AA) \citep[e.g.,][]{seroussi2011authorship, ferracane-etal-2017-leveraging,sharma2018investigation, uchendu-etal-2020-authorship}.

In this study, we utilize the M4 dataset \citep{wang-etal-2024-m4}, which includes texts aligned with human writings and generated by five different models: ChatGPT, Cohere, GPT-
3.5 (text-davinci-003), BLOOMz-176B \citep{muennighoff-etal-2023-crosslingual}, and Dolly-v2 \citet{conover2023free}. The prevalent approach in the field involves applying large language models (LLMs) for classification tasks to distinguish between human and machine-generated texts. Despite their state-of-the-art performance, these models often lack explainability from a human perspective.

Our research aims to bridge this gap by examining the linguistic features of aligned texts generated by humans and LLMs using interpretable features, aiming to find specific features that strongly distinguish machine-generated text and human written texts. 
We profile all texts with the Linguistic Feature Toolkit (LFTK) \citep{lee-lee-2023-lftk}, extracting 247 linguistic features. Furthermore, we add measures for syntactic depth, semantic distance, and emotional content to complement the linguistic profile of each document.

We calculate the average syntactic depth of each documents to examine differences in cognitive demand between documents by using the dependency parser provided by python library Spacy\footnote{\url{https://spacy.io/models/en}}. We calculate the mean and standard error for each model, demonstrating that humans tend to employ more shallow syntactic structure in comparison to generative models.

We conduct analysis at the semantic level to examine differences in semantic consistency between documents by performing pairwise sentence comparisons using cosine similarity and a sentence transformer, specifically the paraphrase-MiniLM-L6-v2 transformer \citep{reimers-2019-sentence-bert}. We calculate the mean and standard error for each model, demonstrating that humans tend to employ less similar semantic content, despite having on average a richer vocabulary. Thus, humans exhibit a lower type-token ratio compared to LLMs. 

Lastly, we investigate the emotionality of the texts to examine differences in the emotional content between documents, hypothesizing that human texts contain a higher degree of emotional expression than machine-generated texts (MGT). Our hypothesis is confirmed, particularly in the presence of negative emotions such as anger, which are more prevalent in human texts.

To better understand the distribution of features for different documents, we finally apply a two-dimensional principal component analysis (PCA) reduction. Results show differences between texts generated by humans and LLMs, especially on the high variability existing in-between human written texts. 

Using a logistic classifier on our corpus, we demonstrate that it is possible to achieve high accuracy (0.82) with human-explainable features. Furthermore, we identify the features that most influence our model's decision for each type of text.

Our main contributions are: (i) we provide an analysis of linguistics components' differences between human-written texts (HWT) and machine-generated texts (MGT) considering different domains, such as  informal online discourse (Reddit) and formal and general knowledge (Wikipedia), (ii) we provide a more systematic analysis by adding multiple features, such as semantic similarity and emotionality, and (iii) we examine differences between texts written by humans and those generated by very recent large language models, such as Cohere and ChatGPT.

\section{Related Work}
Scholars have explored various approaches to tackle the challenge of distinguishing between human-written and machine-generated texts. This task, often referred to as Authorship Attribution (AA), involves detecting whether a text is produced by a human or a generative language model, or attributing authorship among different models.

The attribution of authorship to a text carries significant social relevance, especially in areas such as fake news detection \citep{kumarage2023stylometric, jawahar-etal-2020-automatic}. The need for explainability becomes particularly important when engaging with a broad audience of non-experts, who may not have the means to access or comprehend detection models \citep{gehrmann-etal-2019-gltr}. As a result, numerous studies have focused on identifying human-explainable features that can differentiate between machine-generated (MGT) and human-written text (HWT) \citep[e.g.,][]{dugan2023real, guo2023close, kumarage2023stylometric, uchendu-etal-2020-authorship}. To achieve this, researchers have employed diverse analytical approaches, including stylometric analysis \citep{kumarage2023stylometric, ma2023ai}, qualitative assessments \citep{guo2023close,gehrmann-etal-2019-gltr}, and linguistic feature analysis \citep{wang-etal-2024-m4, uchendu-etal-2020-authorship, ferracane-etal-2017-leveraging}, to diverse corpora, contexts, and generation tasks.

Moreover, classical machine learning algorithms, such as logistic regression, have been employed to train models on bag-of-words features to differentiate between HWT and MGT \citep{solaiman2019release, ippolito-etal-2020-automatic}. Other traditional methods leverage explainable features, including POS-tags \citep{ferracane-etal-2017-leveraging}, topic modeling \citep{seroussi2014authorship}, and LIWC (Linguistic Inquiry and Word Count) features to provide deeper insights into the characteristics of MGT \citep{uchendu-etal-2020-authorship, li-etal-2014-towards}. 

With the advent of Large Language Models (LLMs), detection mechanisms have evolved. Fine-tuned models such as RoBERTa have achieved state-of-the-art performance in many tasks \citep{crothers2023machine, jawahar-etal-2020-automatic}. More recently, zero-shot and few-shot models based on large datasets of HWT and MGT have emerged, offering improved detection capabilities \citep{mitchell2023detectgpt}.

\section{Data}
We use the dataset from Subtask B of SemEval 2024 Task 8 \citep{semeval2024task8} for our study due to its alignment between HWT and MGT from five different LLMs: ChatGPT, Cohere, GPT-3.5 (text-davinci-003), BLOOMz-176B and Dolly-v2. We focus on the subset of the dataset containing texts written in the English language. The corpus comprises a total of 71,027 texts, with approximately 3,000 texts per model and human author in each domain. The domains in the corpus include: formal and general knowledge (Wikipedia), instructional content (Wikihow), formal and scientific writing (Arxiv), and informal (conversational) online discourse (Reddit). 
The machine-generated texts were produced using prompts specifically crafted to align with the content and style of each domain. Each prompt is designed to elicit contextually relevant responses, such as generating a Wikipedia article given its title or writing an abstract for a scientific article based on a given title. In this setting, the outputs of the LLMs are thematically aligned with the corresponding human-written texts, facilitating direct comparisons. The dataset is curated by filtering incoherent or irrelevant outputs, resulting in a high-quality resource that reflect diverse writing styles thus allowing for evaluation of alignment between human and machine-generated texts.

\section{Methodology}

We employ the LFTK tool \citep{lee-lee-2023-lftk} to profile each text for linguistic features. Additionally, we calculate syntactic depth, semantic distance, and emotionality to examine differences in cognitive demand, semantic consistency, and emotional content, respectively. 
We apply a two-dimensional PCA reduction to the features in order to examine differences between humans and language models. Finally, we use a logistic classifier on all calculated features to identify the author of a given text using explainable features. All codes will be released in the camera-ready version due to anonymity reasons.

\subsection{Features Collection}
We present here in detail the linguistic features that we calculate for profiling the human-written (HWT) and machine-generated texts (MGT).\\
\textbf{LFTK's features}: LFTK's features are 247 features measured using the "LFTK" (linguistic feature toolkit) \citep{lee-lee-2023-lftk}. These features are grouped into the following categories: surface, lexico-semantics, discourse and syntax\footnote{A detailed list of the LFTK's features can be found at \url{https://github.com/brucewlee/lftk}}.   \\
\textbf{Syntactic Depth}: Syntactic Depth is calculated using the dependency parser in SpaCy "en\_core\_web\_sm". This feature is included to describe potential differences between HWT and MGT in terms of cognitive demand \citep{hagoort1999neurocognition}.  \\
\textbf{Semantic Distance}: Semantic Distance is calculated using sentence embeddings derived from the Sentence Transformer model "paraphrase-MiniLM-L6-v2". We calculate the cosine similarity in pair-wise sentence comparisons and averaged the distance for each document. This feature describes the semantic content of a text in terms of consistency, as in \citet{beaty2021automating}. \\
\textbf{Emotionality}: Emotional content is calculated using the NRC Emotion Intensity Lexicon \citep{Mohammad13}. The lexicon includes 8 different emotions: anger, disgust, fear, sadness, joy, anticipation,  surprise, trust. We consider anger, fear and sadness as negative emotions, while we take joy as the only representative positive emotion in the lexicon following \citet{aroyehun2023leia}. This feature helps to describe potential differences in emotional content between HWT and MGT, as in \citet{guo2023close}.

\subsection{Analysis}
In our analysis, we examine a comprehensive set of 250 linguistic features across the texts generated by LLMs and those written by humans.
First, we analyze the basic textual property, the length of the texts.

Next, we dive into specific features that have theoretical relevance as meaningful indicators of textual differences. These include the number of unique words, the average syntactic depth, the semantic distance, and the emotional content of the documents. We calculate these features for each LLMs and the human-written texts, and report the average values. We assess the significance of differences in features between humans and LLMs using the Kruskal-Wallis test \citep{mckight2010kruskal}. To further explore pairwise differences, we applied Dunn’s test for multiple comparisons \citep{dinno2015nonparametric}.

To further understand the variability within and between the text sources, we conduct two additional analyses. First, we calculate the variance in the previously mentioned features for each model and the human-written texts, and visualize the results. This allows us to compare the spread of the feature values, providing insights into the cognitive and stylistic differences. Second, we perform a two-dimensional Principal Component Analysis (PCA) on the full set of 250 linguistic features. This dimensionality reduction technique allows us to visualize the overall separation between the text sources in a lower-dimensional space. We then quantify the variability within and between the machine and human-written text clusters. 

Finally, we train a logistic classifier on the full set of linguistic features, following the same-generator, same-domain evaluation protocol as in the SemEval 2024 Task 8 Subtask B \citep{wang-etal-2024-m4}. This allows us to assess the predictive power of the linguistic features in distinguishing between machine-generated and human-written texts.

\section{Results}
The analysis of the corpus reveals significant differences in text length between human-written (HWT) and machine-generated texts (MGT), with humans generally producing longer texts. As shown in Table \ref{table:tokens_per_document}, humans tend to write nearly twice as many words as the models. Consequently, features such as the number of sentences, words, characters, and stop-words are substantially higher in human-written texts.

\begin{table}[h!]
\centering
\begin{tabular}{l|r}
\toprule
\textbf{Model} & \textbf{Tokens per Document} \\
\midrule
bloomz  & 187.31\phantom{000000} \\
chatGPT & 412.03\phantom{000000}  \\
cohere  & 316.83\phantom{000000} \\
davinci & 389.24\phantom{000000} \\
dolly   & 373.97\phantom{000000}  \\
human   & 706.16\phantom{000000} \\
\bottomrule
\end{tabular}
\caption{Mean Tokens per Document for LLMs and Humans}
\label{table:tokens_per_document}
\end{table}

We assess the number of unique words used by humans and LLMs, showing a broader vocabulary and greater variability among human authors. MGT, on the other hand, are more concise and consistent (See Figure \ref{fig:average_average_unique_words} and Figure \ref{fig:boxplot_by_model_uniquewords} in Appendix).

\begin{figure}[h]
    \centering
    \includegraphics[width=\linewidth]{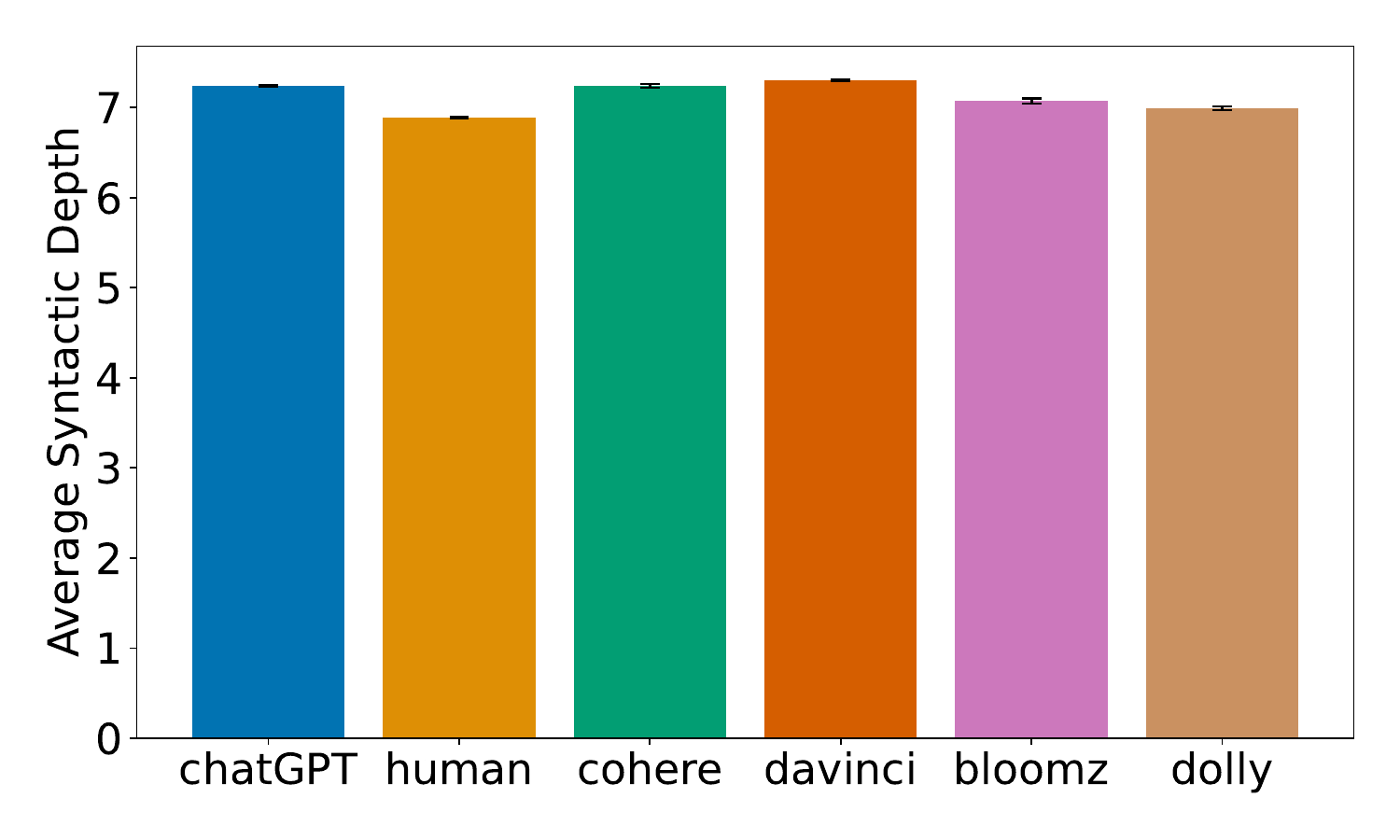}
    \caption{Average Syntactic Depth for LLMs and Humans}
    \label{fig:average_syntactic_depth_general}
\end{figure}

\begin{figure}[h]
    \centering
    \includegraphics[width=\linewidth]{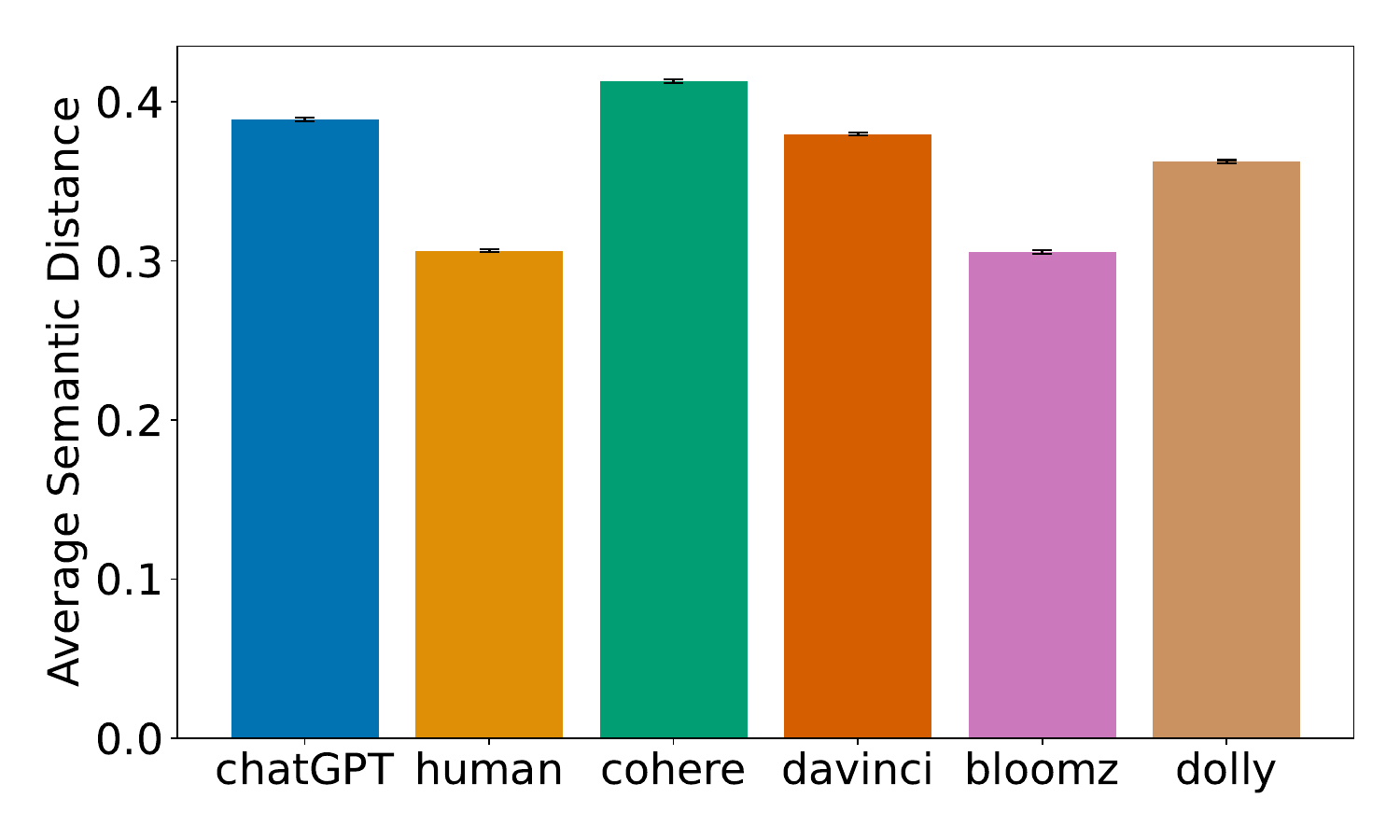}
    \caption{Semantic Distance pair-wise Sentence comparison for LLMs and Humans}
    \label{fig:semantic_distance}
\end{figure}

Despite the difference in text length, humans score significantly lower in syntactic depth compared to LLMs, as illustrated in Figure \ref{fig:average_syntactic_depth_general}. Literature on the cognitive demands of syntactic structures suggests that humans prefer less complex sentences to manage cognitive load \citep{hagoort1999neurocognition}.
The variability in the number of unique words among HWT is linked to their low semantic similarity, as shown in Figure \ref{fig:semantic_distance}. However, this does not fully account for the consistency observed in HWT, where measurements of variability in Figure \ref{fig:variance_semantic_distance} in the Appendix indicates that humans generate more meaningful and diverse content consistently.

\begin{figure}[h]
    \centering
    \includegraphics[width=\linewidth]{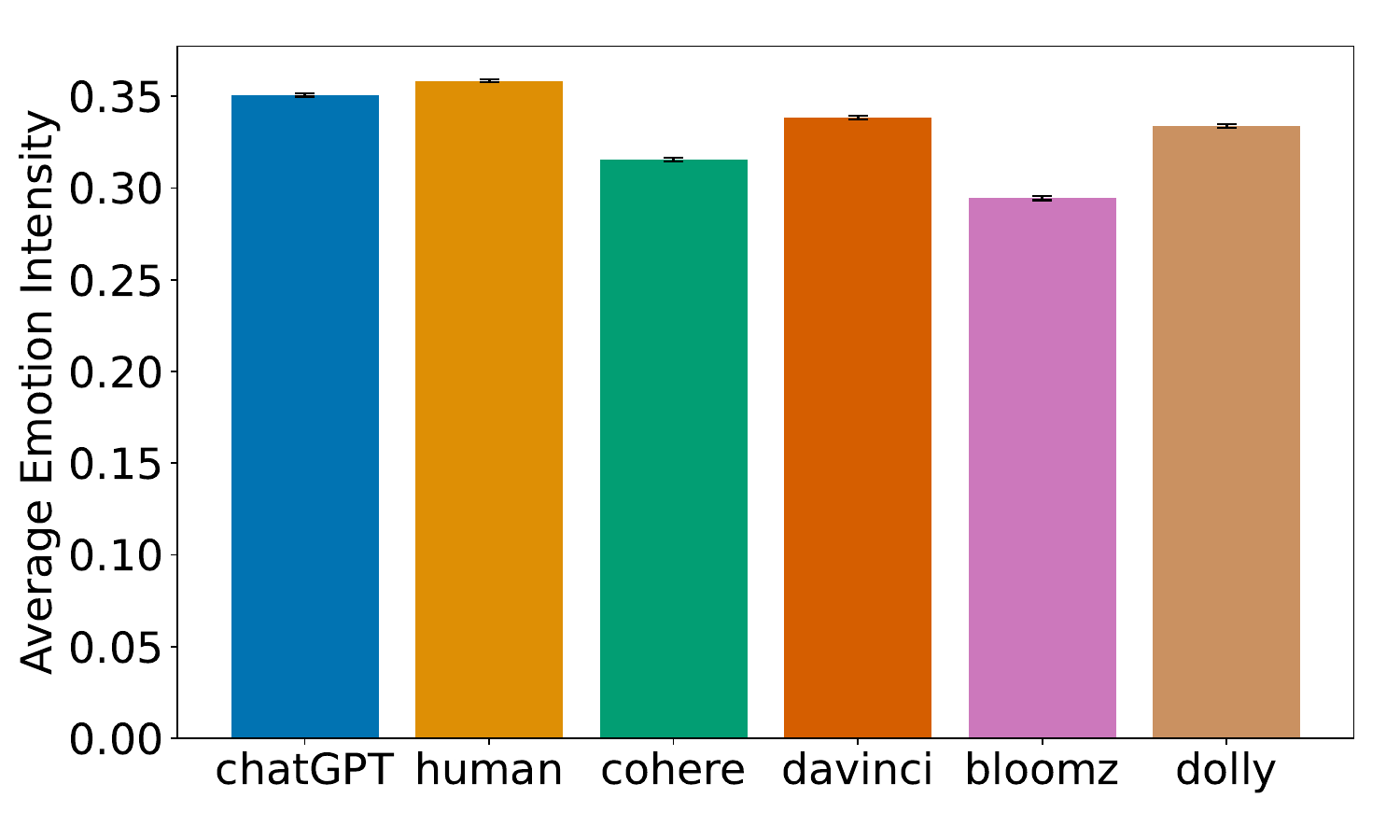}
    \caption{Average Emotion Intensity for LLMs and Humans}
    \label{fig:average_emotion_intensity_general}
\end{figure}

\begin{figure}[h]
    \centering
    \includegraphics[width=\linewidth]{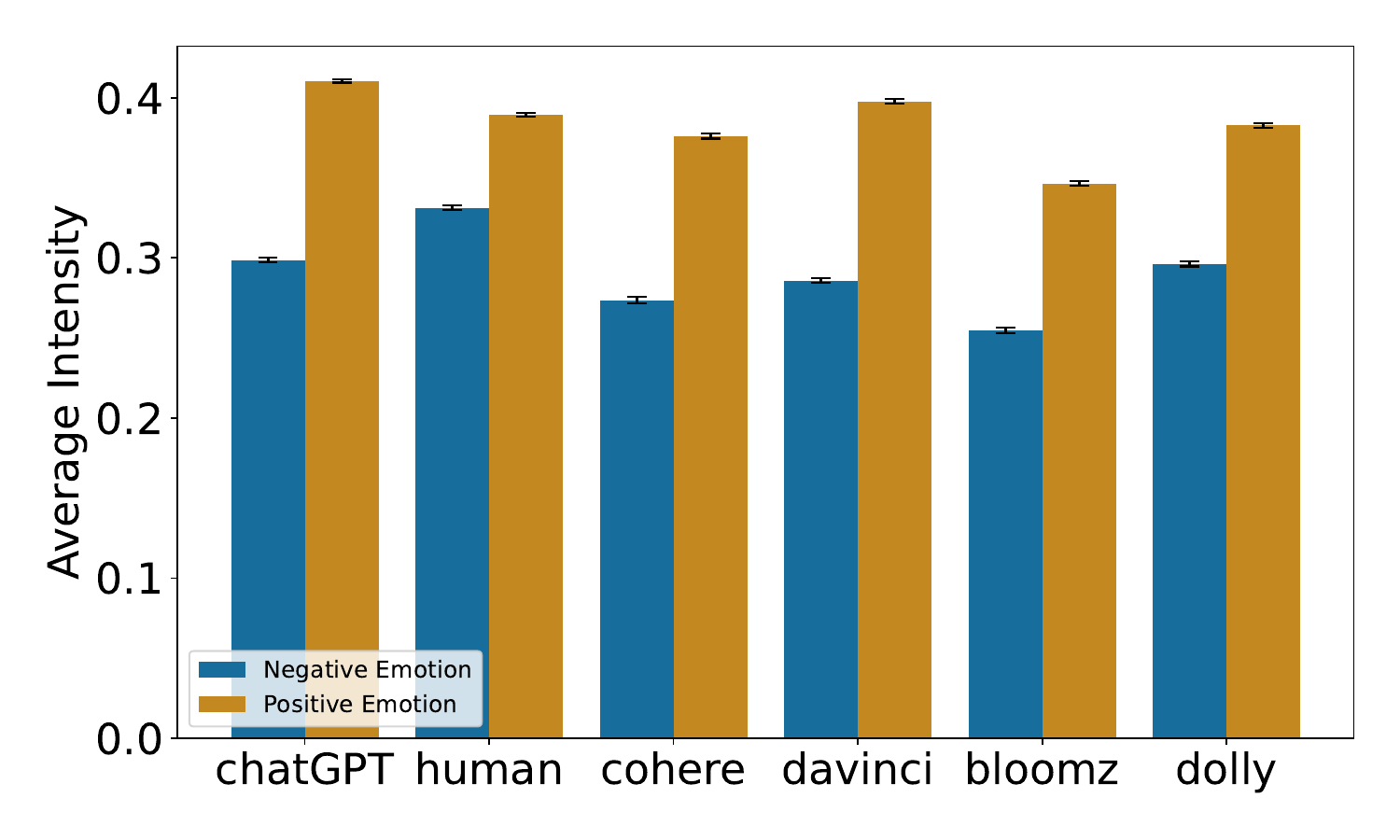}
    \caption{Average Negative and Positive Emotions for LLMs and Humans}
    \label{fig:average_emotion_intensity_per_source}
\end{figure}

A similar trend is observed in the analysis of emotional content in Figure \ref{fig:average_emotion_intensity_general}, where HWT exhibit higher emotionality, especially negative emotions, as can be seen in Figure \ref{fig:average_emotion_intensity_per_source}. These findings align with prior research suggesting that MGT often show a narrower spectrum of emotions, downplaying negative emotions \citep{guo2023close}.

Table \ref{KW} shows the statistically significant differences in features between LLMs and between humans and LLMs, as confirmed by the Kruskal-Wallis test. Additionally, Table \ref{dunn} in the Appendix provides pairwise comparisons between LLMs and humans, conducted using Dunn's test. For example, we observe how ChatGPT and davinci score similarly in \texttt{Unique Words}. This similarity is likely attributable to their shared origin from the same provider (OpenAI) and their close proximity in model versions. 

\begin{table}[h]
\centering
\begin{tabular}{l|r|r}
\toprule
Feature & H-statistic & p-value \\
\midrule
Semantic Distance & 8548.76 & <0.001 \\
Unique Words               & 11575.71 & <0.001 \\
Syntactic Depth            & 1246.65 & <0.001 \\
Average Emotion            & 2402.78 & <0.001 \\
\bottomrule
\end{tabular}
\caption{Kruskal-Wallis Test Results}
\label{KW}
\end{table}

\begin{figure}[h]
    \centering
    \includegraphics[width=\linewidth]{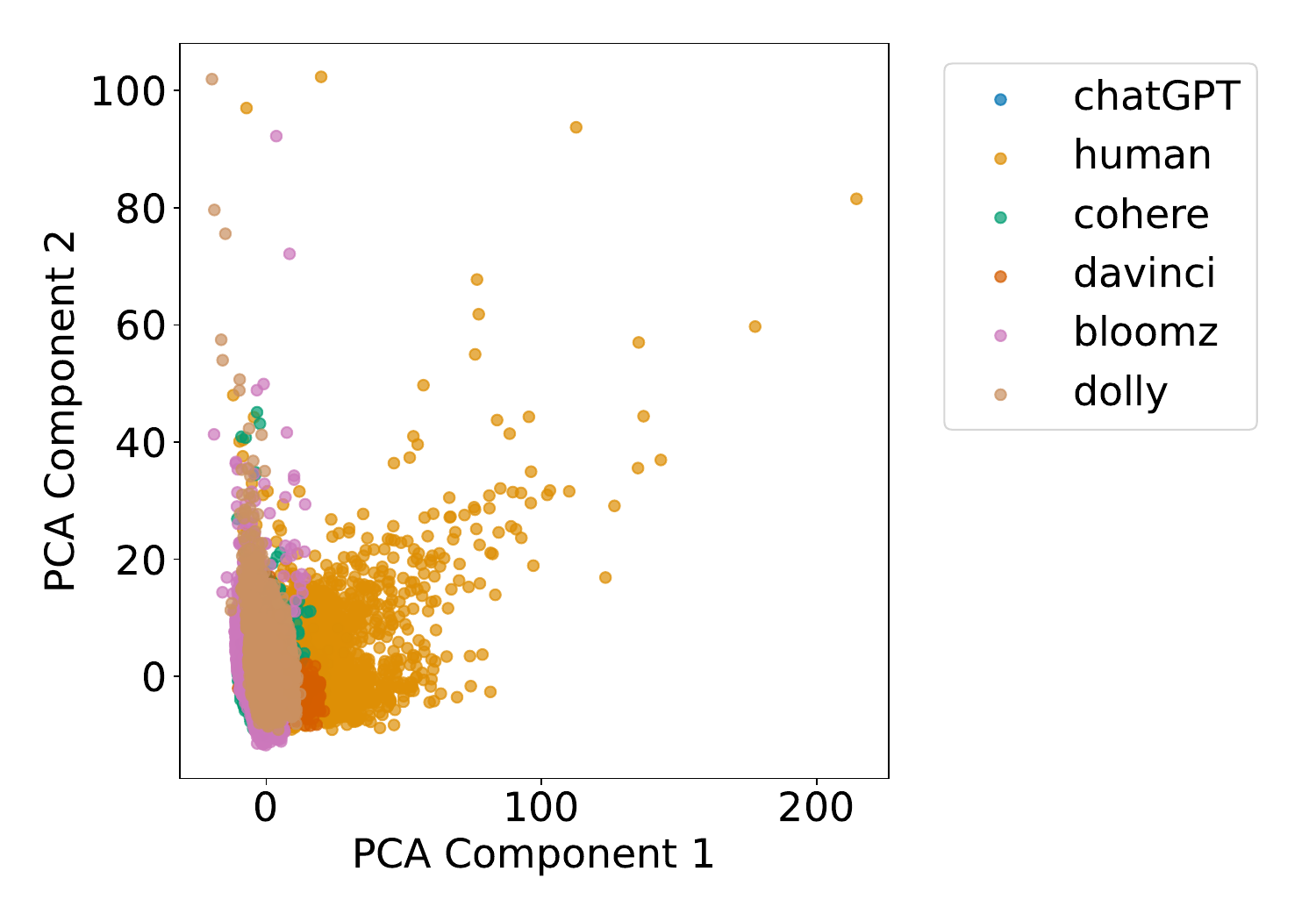}
    \caption{Component Reduction of Text Features for LLMs and Humans}
    \label{fig:text_features_pca}
\end{figure}

Figure \ref{fig:text_features_pca} shows the PCA results for all linguistic features. The LLMs are difficult to distinguish because they show very similar scores across various linguistic features. Component 1 explains 18.82\% of the variance in the data, while Component 2 explains 8.92\%. The variability in texts is calculated by measuring the distance from their centroids in the two-dimensional PCA space. Table \ref{table:scores_by_model_and_source} indicates that HWT show highly diverse linguistic profiles, particularly across different domains with specific stylistic constraints \citep{biber2019register}. 
Humans exhibit greater linguistic variation in domains where the constraints on language use are less rigid, such as Wikipedia or Reddit. In contrast, humans score more similarly to the LLMs in domains with stricter linguistic constraints, such as ArXiv.

\begin{table}[h]
\centering
\footnotesize
\begin{tabular}{l|c|c|c|c}
\toprule
\textbf{Model} & \textbf{WikiHow} & \textbf{Wikipedia} & \textbf{Reddit} & \textbf{arXiv} \\
\midrule
chatGPT & 17.71 & 4.72 & 7.70 & 1.64 \\
human   & 167.20 & 296.75 & 50.67 & 4.70 \\
cohere  & 13.35 & 16.71 & 12.65 & 4.30 \\
davinci & 23.66 & 6.47 & 6.60 & 4.19 \\
bloomz  & 25.00 & 25.66 & 14.23 & 9.74 \\
dolly   & 33.69 & 34.57 & 12.94 & 9.10 \\
\bottomrule
\end{tabular}
\caption{Variability of PCA scores by Domain for LLMs and Humans}
\label{table:scores_by_model_and_source}
\end{table}

While the performance of the logistic classifier did not reach that of other models reported in \citet{wang-etal-2024-m4}, it 
still achieved  an accuracy of 0.87 (See Table \ref{table:classification_results} in the Appendix). 
Additionally, as shown in Table \ref{table:top_features_by_model} in the Appendix, the logistic classifier’s explainable features indicate that the number of words and unique words are the most prominent variables for predicting HWT, achieving a prediction accuracy of 93\%.

\section{Discussion} \label{p:dis}

Our analysis of linguistic features highlights differences between human-written (HWT) and machine-generated texts (MGT) in the target corpus. The first distinction is the length of the texts: humans tend to write longer and more varied texts, whereas LLMs produce shorter and more structured content \citep{info14100522, BARROT2023100745}. Moreover, our PCA analysis shows that human-written texts exhibit greater variability depending on the context and domain \citep{guo2023close}. This suggests a strong influence of individual style where linguistic constraints are flexible, raising questions about the homogenization of linguistic features when generated by LLMs that are extremely more consistent than humans \citep{guo2023close}.

Despite these individual differences, humans score lower in syntactic depth compared to LLMs. This reflects the cognitive demands that humans experience when constructing sentences \citep{hagoort1999neurocognition}. While LLMs can emulate HWT effectively, they do not face the same cognitive constraints and thus tend to produce more syntactically complex sentence.

Furthermore, humans exhibit a richer vocabulary than LLMs \citep{wang-etal-2024-m4, guo2023close}, aligning with our semantic similarity results. HWT contain richer semantic content and show greater consistency in meaning than LLMs. Additionally, HWT are richer in emotional content, particularly negative emotions \citep{guo2023close}. This suggests that the training procedure of LLMs such as through alignment may have led to the reduction of negative emotional expressions, in a bid to make LLMs more helpful and less harmful.

Finally, the classification task indicates that we can reach good performances in identifying the authorship of texts with a logistic classifier trained solely on linguistic features. Our feature analysis reveals that the logistic classifier often relies on different features to classify HWT and MGT (Refer to Section \ref{classfeat} in the Appendix). Indeed, humans tend to be less consistent and formal in structuring texts than LLMs \citep{info14100522, BARROT2023100745, guo2023close}. For example, an analysis of the features driving our logistic classification shows the top features in classifying HWT to be number of unique words and number of unique adverbs. In contrast, the classification of texts generated by many LLMs rely on number of spaces or reading time.
Despite the qualitative nature of this analysis, our findings underscore the necessity for developing more human-explainable detection tools to better understand and identify outputs of LLMs.

\section{Conclusion and Future Work}
In our analysis of linguistic components of aligned human-written (HWT) and machine-generated texts (MGT), we identified several key differences that characterize our target corpus. We show that humans show extremely varied linguistic profiles whenever the linguistic domain imposes less rigid constraints, whereas LLMs vary sensitively less in this regard. We show that human texts are richer in vocabulary and longer than those produced by LLMs. However, human texts tend to be less syntactically complex, reflecting the cognitive load required to process them. Additionally, HWT are more semantically diverse and consistent, and they exhibit higher levels of emotional content, particularly negative emotions. Our findings underline the importance of incorporating meaningful features from a human perspective to improve the understanding of differences between human-written and machine-generated texts.

Future work should explore diverse corpora with varying characteristics to verify these differences across different domains and languages. Expanding the range of linguistic features, especially those related to content such as the use of metaphors or figurative language, could provide deeper insights. To improve the accuracy of classification tasks using linguistic features alone, non-linear models like random forests could be applied to human-machine authorship attribution tasks.

\section{Limitations}
One limitation of our study lies in its generalizability. We rely on a dataset covering four domains in the English language. Thus, the applicability of our findings is limited when considering broader linguistic variations across domains and languages. The four domains covered in our dataset provide valuable insights, but they may not be representative of all possible linguistic contexts characteristic of all textual domains. Furthermore, texts in various languages can have unique linguistic features which limits the relevance of our results to non-English contexts.

Our analysis focuses on a set of LLMs which may already have been superseded by more advanced versions due to the rapid advancements in the field. This poses a challenge for the temporal validity of our findings as future LLMs could exhibit different linguistic patterns. 
However, this limitation also highlights an interesting opportunity to study the evolution of linguistic patterns in textual outputs of LLMs over time. Such a study could potentially reveal trends in how linguistic characteristics of text generated by LLMs align with technological advancements.

Extending this study to multiple languages, datasets, and model versions could potentially enhance the applicability of our findings across broader contexts and evolving technologies. However, such an extension would require significant data curation efforts and depends on the availability of multilingual linguistic pipelines capable of profiling texts at scale. Nevertheless, this study and our findings can serve as a foundation for further exploration of LLM outputs.

\section{Ethical Considerations}
In developing our approach, we acknowledge the potential for unintended bias, particularly against non-native English speakers. Some of the linguistic features we analyze may capture characteristics in texts produced by English language learners. This overlap raises important ethical questions.
It is crucial to emphasize that our primary objective is to advance the theoretical understanding of language patterns in texts generated by humans and LLMs, rather than to create tools for real-world applications. The features and techniques described in this paper are intended for research purposes and should not be directly applied in practical systems without careful consideration of their broader implications. Any potential real-world application would require extensive additional research and safeguards. We strongly caution against using these features or similar approaches in high-stakes decision-making processes or in any context where they could disadvantage individuals based on their language competency.

\bibliography{custom}

\section{Appendix}

\begin{figure}[ht]
    \centering
    \includegraphics[width=\linewidth]{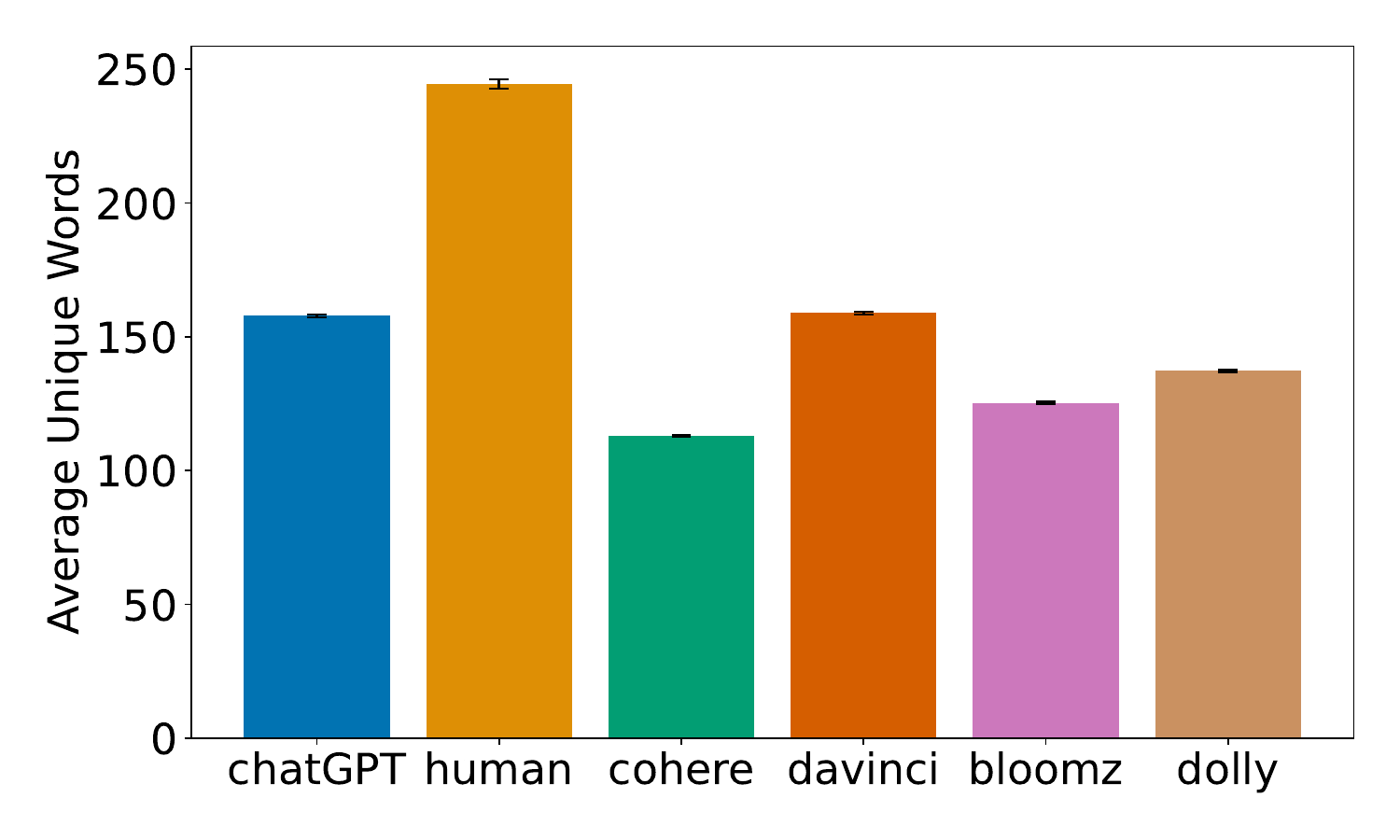}
    \caption{Average Unique Words for LLMs and Humans}
    \label{fig:average_average_unique_words}
\end{figure}

\begin{figure}[ht]
    \centering
    \includegraphics[width=\linewidth]{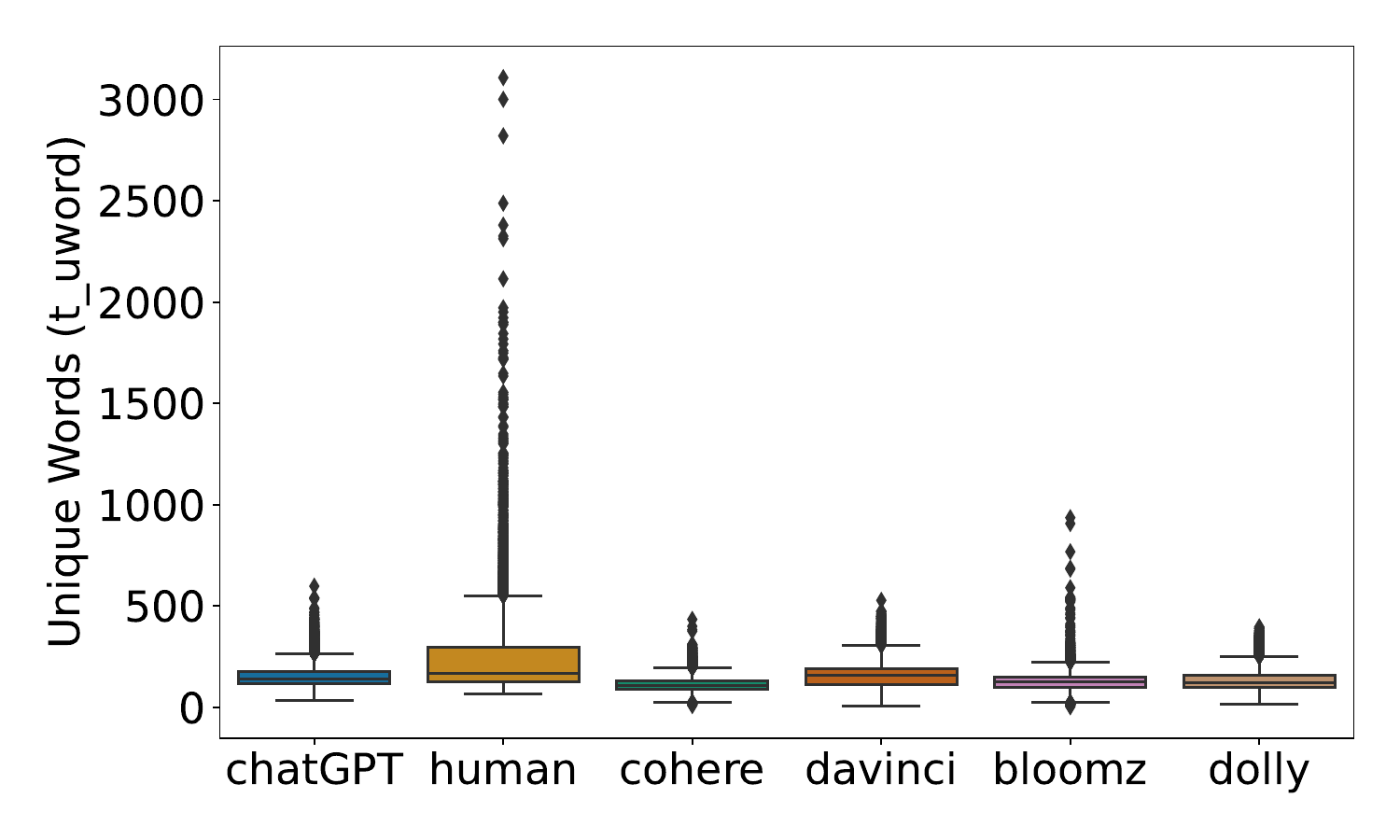}
    \caption{Variability of Unique Words for LLMs and Humans}
    \label{fig:boxplot_by_model_uniquewords}
\end{figure}

In the Appendix, we present an analysis of the differences between human-written (HWT) and machine-generated text (MGT). Figure \ref{fig:average_average_unique_words} reports the differences in the number of unique words used by different models, showing a richer vocabulary in HWT. Moreover, Figure \ref{fig:boxplot_by_model_uniquewords} shows the variety of the number of unique terms used in each document, where HWT exhibit significantly greater variability compared to MGT. 

Table \ref{dunn} shows the statistically significant differences in features between LLMs and between humans and LLMs in a pairwise comparisons, conducted using Dunn's test. A value below 0.05 indicates a statistically significant difference between the corresponding LLMs and/or human.

We further explore these differences by detailing an in-domain, same-generator classification task. In this context, "in-domain" refers to the inclusion of all relevant domains in the training set, while "same-generator" indicates that all models used in the analysis are also present in the training set (see Section \ref{classfeat} for more details).

\begin{figure}[ht]
    \centering
    \includegraphics[width=\linewidth]{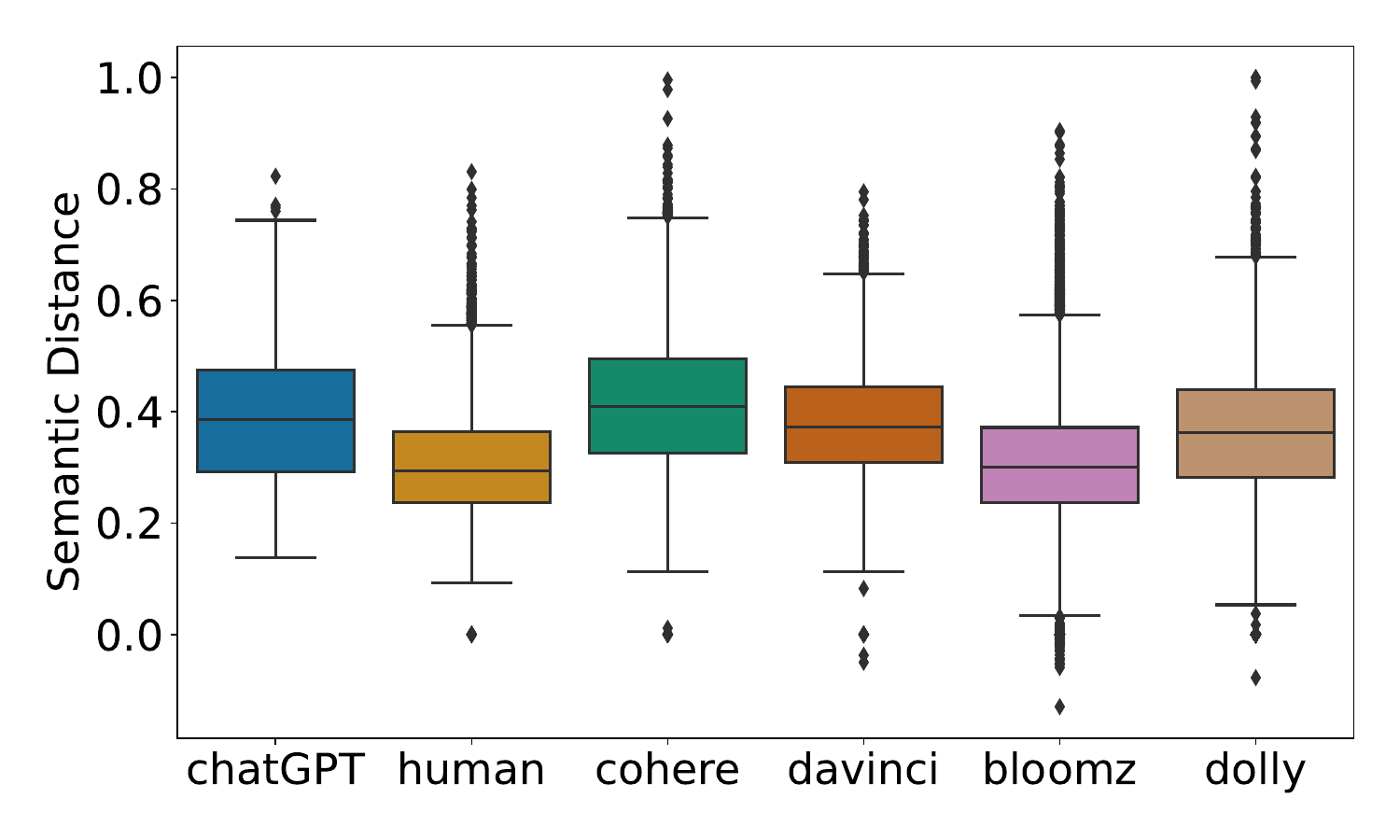}
    \caption{Variability of Semantic Distance pair-wise Sentence comparison for LLMs and Humans}
    \label{fig:variance_semantic_distance}
\end{figure}

\begin{table*}[htbp]
\centering
\caption{Dunn's Test Results for Different Features Across Models}
\label{dunn}
\begin{tabular}{l|r|r|r|r|r|r}
\toprule
Feature & bloomz & chatGPT & cohere & davinci & dolly & human \\
\midrule
\multicolumn{7}{c}{\textbf{Semantic Distance}} \\
bloomz  &  1.000 &  <0.001 &   <0.001 &  <0.001 &   <0.001 &  <0.001 \\
chatGPT &  <0.001 &  1.000 &   <0.001 &  0.020 &   <0.001 &  <0.001 \\
cohere  &  <0.001 &  <0.001 &   1.000 &  <0.001 &  <0.001 &  <0.001 \\
davinci &  <0.001 &  0.020 &   <0.001 &  1.000 &   <0.001 &  <0.001 \\
dolly   &  <0.001 &  <0.001 &  <0.001 &  <0.001 &   1.000 &  <0.001 \\
human   &  <0.001 &  <0.001 &   <0.001 &  <0.001 &   <0.001 &  1.000 \\
\midrule
\multicolumn{7}{c}{\textbf{Unique Words}} \\
bloomz  &   1.000 &  <0.001 &  <0.001 &   <0.001 &   <0.001 &   <0.001 \\
chatGPT &  <0.001 &   1.000 &   <0.001 &   0.160 &  <0.001 &  <0.001 \\
cohere  &  <0.001 &   <0.001 &   1.000 &   <0.001 &  <0.001 &   <0.001 \\
davinci &   <0.001 &   0.160 &   <0.001 &   1.000 &  <0.001 &  <0.001 \\
dolly   &   <0.001 &  <0.001 &  <0.001 &  <0.001 &   1.000 &   <0.001 \\
human   &   <0.001 &  <0.001 &   <0.001 &  <0.001 &   <0.001 &   1.000 \\
\midrule
\multicolumn{7}{c}{\textbf{Syntactic Depth}} \\
bloomz  &   1.000 &   <0.001 &  <0.001 &  <0.001 &  0.030 &   1.000 \\
chatGPT &   <0.001 &   1.000 &  <0.001 &   0.010 &  <0.001 &  <0.001 \\
cohere  &   <0.001 &   <0.001 &  1.000 &   <0.001 &  <0.001 &   <0.001 \\
davinci &  <0.001 &   0.010 &  <0.001 &   1.000 &  <0.001 &  <0.001 \\
dolly   &   0.030 &   <0.001 &  <0.001 &   <0.001 &  1.000 &   <0.001 \\
human   &   1.000 &  <0.001 &  <0.001 &  <0.001 &  <0.001 &   1.000 \\
\midrule
\multicolumn{7}{c}{\textbf{Average Emotion}} \\
bloomz  &   1.000 &  <0.001 &  <0.001 &  <0.001 &  <0.001 &   <0.001 \\
chatGPT &  <0.001 &   1.000 &  <0.001 &   <0.001 &   <0.001 &   <0.001 \\
cohere  &   <0.001 &  <0.001 &   1.000 &  <0.001 &   <0.001 &  <0.001 \\
davinci &  <0.001 &   <0.001 &   <0.001 &   1.000 &   <0.001 &   <0.001 \\
dolly   &  <0.001 &   <0.001 &   <0.001 &   <0.001 &   1.000 &   <0.001 \\
human   &   <0.001 &   <0.001 &  <0.001 &   <0.001 &   <0.001 &   1.000 \\
\bottomrule
\end{tabular}
\end{table*}

\subsection{In-domain Same-generator Classification with Logistic Regression} \label{classfeat}
In this analysis, we trained a logistic regression model to classify text from humans and five different LLMs: ChatGPT, Cohere, GPT-3.5 (text-davinci-003), BLOOMz-176B and Dolly-v2. The training set consisted of the remaining data after randomly selecting 5,000 elements from each model for the test set (Refer to Table \ref{table:dataset_compositions} for a detailed view on the train-test split). After training, the model was evaluated on the 30,000-element test set, yielding an overall accuracy of 87.15\%. The classification performance, detailed in Figure \ref{table:classification_results}, shows that precision, recall, and F1-scores vary across models, with BLOOMz achieving the highest individual metrics.

Furthermore, we explored the features that most influence the logistic classifier for each model. We identified the top 10 features by coefficient for each model, as presented in table \ref{table:top_features_by_model}. Each row represents a feature, and the cells contain the feature's coefficient for the corresponding model, with a \textit{-} indicating that the feature is not among the top 10 for that particular model. The features that characterize the human model differ significantly from those of other models. For instance, while most LLMs like BLOOMz, ChatGPT, and Dolly frequently emphasize structural features like \texttt{n\_uspace} (number of unique spaces, \texttt{n\_sconj} (number of subordinating conjunctions), and \texttt{cole} (reading time with the coleman liau index \citep{liau1976modification}), humans are more influenced by features related to linguistic richness and diversity, such as \texttt{t\_uword} (total unique words), \texttt{a\_propn\_pw} (average proper nouns per word), and \texttt{n\_noun} (number of nouns).

In contrast, LLMs often rely on structural features, reflecting the more formal nature of their generated content. This distinction underscores the differences in how humans and LLMs produce language, with humans exhibiting a broader range of linguistic profiles and LLMs favoring consistency and specific patterns in text generation.

\begin{table*}[ht!]
\centering
\begin{tabular}{l|c|c}
\toprule
\textbf{Model} & \textbf{Training Set Composition} & \textbf{Test Set Composition} \\
\midrule \hline
chatGPT & 6995 & 5000 \\
bloomz  & 6998 & 5000 \\
dolly   & 6702 & 5000 \\
human   & 6997 & 5000 \\
cohere  & 6336 & 5000 \\
davinci & 6999 & 5000 \\
\bottomrule
\end{tabular}
\caption{Training and Test Set Compositions for LLMs and Humans}
\label{table:dataset_compositions}
\end{table*}

\begin{table*}[ht!]
\centering
\begin{tabular}{l|c|c|c}
\toprule
\textbf{Model} & \textbf{Precision} & \textbf{Recall} & \textbf{F1-Score} \\
\midrule \hline
bloomz  & 0.97 & 0.97 & 0.97 \\
chatGPT & 0.79 & 0.79 & 0.79 \\
cohere  & 0.87 & 0.86 & 0.87 \\
davinci & 0.76 & 0.73 & 0.75 \\
dolly   & 0.90 & 0.94 & 0.92 \\
human   & 0.92 & 0.93 & 0.93 \\
\midrule \hline
\textbf{Overall Accuracy} & \multicolumn{3}{c}{0.87
} \\
\bottomrule
\end{tabular}
\caption{Classification Results for LLMs and Humans}
\label{table:classification_results}
\end{table*}

\begin{table*}[ht!]
\centering
\footnotesize
\begin{tabular}{l|c|c|c|c|c|c}
\hline
\textbf{Features} & \textbf{bloomz} & \textbf{chatGPT} & \textbf{cohere} & \textbf{davinci} & \textbf{dolly} & \textbf{human} \\
\hline
total\_number\_of\_unique\_spaces & 4.42 & 3.65 & -    & -    & 3.25 & -    \\
total\_number\_of\_subordinating\_conjunctions & 2.82 & -    & -    & -    & -    & -    \\
total\_number\_of\_unique\_words & 2.66 & -    & -    & -    & -    & 3.27 \\
total\_number\_of\_unique\_determiners & 2.58 & -    & 1.54 & -    & -    & -    \\
simple\_type\_token\_ratio\_no\_lemma & 2.52 & 3.22 & -    & 1.67 & -    & -    \\
simple\_type\_token\_ratio & 2.52 & 3.22 & -    & 1.67 & -    & -    \\
total\_number\_of\_unique\_adjectives & 2.46 & 2.47 & -    & 1.68 & -    & -    \\
average\_number\_of\_interjections\_per\_word & 2.13 & -    & -    & 2.39 & -    & -    \\
total\_number\_of\_named\_entities\_date & 2.11 & -    & -    & -    & -    & -    \\
simple\_punctuations\_variation & 2.04 & -    & 1.27 & -    & -    & -    \\
total\_number\_of\_spaces & -    & 3.54 & 1.77 & -    & 3.88 & -    \\
coleman\_liau\_index & -    & 3.41 & 1.55 & -    & 2.80 & -    \\
bilogarithmic\_type\_token\_ratio\_no\_lemma & -    & 2.72 & -    & -    & 2.05 & -    \\
bilogarithmic\_type\_token\_ratio & -    & 2.72 & -    & -    & 2.05 & -    \\
total\_number\_of\_verbs & -    & 2.59 & -    & -    & -    & -    \\
total\_number\_of\_named\_entities\_cardinal & -    & 2.21 & -    & -    & -    & -    \\
average\_subtlex\_us\_zipf\_of\_words\_per\_sentence & - & -    & 1.96 & -    & -    & -    \\
total\_number\_of\_characters & -    & -    & 1.68 & 1.47 & -    & -    \\
average\_subtlex\_us\_zipf\_of\_words\_per\_word & - & -    & 1.53 & -    & -    & -    \\
average\_number\_of\_characters\_per\_sentence & -    & -    & 1.44 & -    & -    & -    \\
average\_number\_of\_verbs\_per\_sentence & -    & -    & 1.35 & -    & -    & -    \\
average\_number\_of\_syllables\_per\_sentence & -    & -    & 1.35 & -    & -    & -    \\
total\_subtlex\_us\_zipf\_of\_words & -   & -    & -    & 3.05 & -    & -    \\
total\_brysbaert\_age\_of\_acquistion\_of\_words & -    & -    & -    & 2.46 & -    & 1.62 \\
simple\_spaces\_variation & -    & -    & -    & 1.97 & -    & -    \\
total\_number\_of\_unique\_nouns & -    & -    & -    & 1.59 & -    & -    \\
corrected\_nouns\_variation & -    & -    & -    & 1.42 & -    & -    \\
corrected\_spaces\_variation & -    & -    & -    & -    & 3.67 & -    \\
root\_spaces\_variation & -    & -    & -    & -    & 3.66 & -    \\
average\_number\_of\_spaces\_per\_word & -    & -    & -    & -    & 3.05 & 1.68 \\
total\_number\_of\_unique\_verbs & -    & -    & -    & -    & 2.50 & -    \\
total\_number\_of\_sentences & -    & -    & -    & -    & -    & 2.53 \\
total\_number\_of\_adpositions & -    & -    & -    & -    & -    & 1.80 \\
average\_number\_of\_proper\_nouns\_per\_word & -    & -    & -    & -    & -    & 1.79 \\
average\_number\_of\_spaces\_per\_sentence & -    & -    & -    & -    & -    & 1.75 \\
simple\_adpositions\_variation & -    & -    & -    & -    & -    & 1.67 \\
\hline
\end{tabular}
\caption{Top 10 Features for LLMs and Humans}
\label{table:top_features_by_model}
\end{table*}

\end{document}